# Soft Biometrics: Gender Recognition from Unconstrained Face Images using Local Feature Descriptor

Olasimbo Ayodeji Arigbabu[1], Sharifah Mumtazah Syed Ahmad[1], Wan Azizun Wan Adnan[1], Salman Yussof [2], Saif Mahmood[1]

[1]Department of Computer and Communication Systems, Universiti Putra Malaysia, Selangor, Malaysia.
[2]Department of Systems and Networking, UniversitiTenaga National, Kajang, Malaysia.

[1]oa.arigbabu@gmail.com,
[1] s_mumtazah@upm.edu.my,
[1]wawa@upm.edu.my,
[2]salman@uniten.edu.my,
[1]saiffd2000@gmail.com

## ABSTRACT

Gender recognition from unconstrained face images is a challenging task due to the high degree of misalignment, pose, expression, and illumination variation. In previous works, the recognition of gender from unconstrained face images is approached by utilizing image alignment, exploiting multiple samples per individual to improve the learning ability of the classifier, or learning gender based on prior knowledge about pose and demographic distributions of the dataset. However, image alignment increases the complexity and time of computation, while the use of multiple samples or having prior knowledge about data distribution is unrealistic in practical applications. This paper presents an approach for gender recognition from unconstrained face images. Our technique exploits the robustness of local feature descriptor to photometric variations to extract the shape description of the 2D face image using a single sample image per individual. The results obtained from experiments on Labeled Faces in the Wild (LFW) dataset describe the effectiveness of the proposed method. The essence of this study is to investigate the most suitable functions and parameter settings for recognizing gender from unconstrained face images.

**Keywords**—Face Gender Recognition, Unconstrained Face Recognition, Soft Biometric Traits, Local Feature Descriptor, Shape Feature Extraction, Kernel Functions, Support Vector Machine

## INTRODUCTION

Over the past few years, recognition of human gender from unconstrained face images has become a recent trend in the research area of facial soft biometrics. This is mainly due to the need of recognizing gender from face images acquired from uncontrolled sources such as online internet web pages, CCTV, webcam, and mobile devices. Besides, due to the fact that gender is an example of soft biometrics (Arigbabu et al., 2014a), it provides detailed descriptions about individuals, which makes gender recognition a useful application in image retrieval, human computer interaction, visual surveillance, commercial market analysis, and improving traditional biometric recognition (Arigbabu et al., 2014; Farinella & Dugelay, 2012). It is worth noting that, gender recognition from constrained frontal face images is already a well-established area of research, as described in several literatures (Arigbabu et al., 2014; Golomb et al., 1990; Khan et al., 2011; Tamura et al., 1996; Rai & Khanna, 2012; Yang & Moghaddam, 2000). Some of the previous works studying face gender recognition problem have utilized multiple sample images per individual to improve the learning ability of the classifiers (Chu et al., 2013; Mayo & Zhang, 2008). While others have exploited image alignment techniques to improve the performance of gender recognition (Makinen & Raisamo, 2008; Shan, 2012). However, the computational complexity of automatic image alignment and the possibility of the optimizer to get trapped in the local minima are amongst the disadvantages of image alignment (Makinen & Raisamo, 2008). In addition, compiling several images of an individual in an unconstrained environment is less practical, because the subjects are not cooperative, and only one or few sample images are available for each individual.

With regard to that, a limited number of studies have been carried out on unconstrained face gender recognition. To this end, the problem has been approached from a variety of perspectives. Some studies have exploited the demographic distribution of the database to aid gender recognition from unconstrained images, such as recognizing gender based on the age of the subjects (Dago-Casas et al., 2011) as shown in Table 1. Some other studies have systematically performed gender recognition based on the pose of the subjects after face detection (Bekios-Calfa et al., 2013; 2014) as highlighted in Table 1. Although, these techniques have shown some level of reliability and robustness, the general assumption is that the data follow the same normal distribution, which is not the case in images acquired from arbitrary sources.

We present in this paper a different approach to evaluate gender recognition. Our assumption is mainly based on the fact that a single image is available for each individual in the database, whereby the construction of the dataset exhibit a large a variation in terms of pose, expression, alignment, occlusion, and illumination conditions. As such, the contribution to knowledge is to transfer the problem of gender to finding a more concise face description. Although, these descriptions are independent for each individual, but they are identically distributed to sufficiently represent the differences between the two genders. The proposed technique does not necessitate image alignment and is fully dependent on the holistic facial features of the individuals to be recognized. Our approach involves extracting face shape description by combining Laplacian filtered images with Pyramid Histogram of Gradient (PHOG) shape descriptor, presented by (Bosch et al., 2007) to aid gender recognition. In addition, we utilized Support Vector Machine (SVM) (Cortes & Vapnik, 1995) with different kernel functions to investigate the face gender recognition problem. The result attained attests to the efficiency of the proposed technique, as the performance is comparable to recent related methods as shown in Table 1. A**n earlier version of this work was presented at ICCOINS 2014** (Arigbabu et al., 2014b). The remainder of the paper is arranged as follows. In Section 2, the proposed system for gender recognition is explained in detail. Section 3 presents the setup of the experiment and the experimental results attained. Finally, discussion and conclusion is provided in Section 4.

**Table 1**

**Comparison of unconstrained face gender recognition techniques**

| Technique | Dataset | Feature Extraction | Classifier | Factors Considered | | | | Result (%) |
| --- | --- | --- | --- | --- | --- | --- | --- | --- |
| | | | | Age | Pose | Alignment | Multiple samples | |
| (Dago-casas et al., 2011) | GROUPS LFW | Raw pixels, Gabor Jets, LBP | SVM LDA | Yes | No | Yes | Yes | 86.4 94.01 |
| (Juan Bekios-Calfa et al., 2014) | LFW/ GROUPS | LDA | KNN | Yes | Yes | No | Yes | 72.01 79.53 |
| (J Bekios-Calfa, 2013) | GROUPS | PCA/LDA | Linear Classifier | Yes | Yes | No | Yes | 81.56 |
| **Proposed** | **LFW** | **Laplacian PHOG** | **SVM** | **No** | **No** | **No** | **No** | **88.5** |

## UNCONSTRAINED FACE GENDER RECOGNITION

The technique proposed in this paper utilizes the already established basic operations for carrying out recognition in biometric applications, which includes pre-processing, feature extraction, and classification, as depicted in Figure 1.

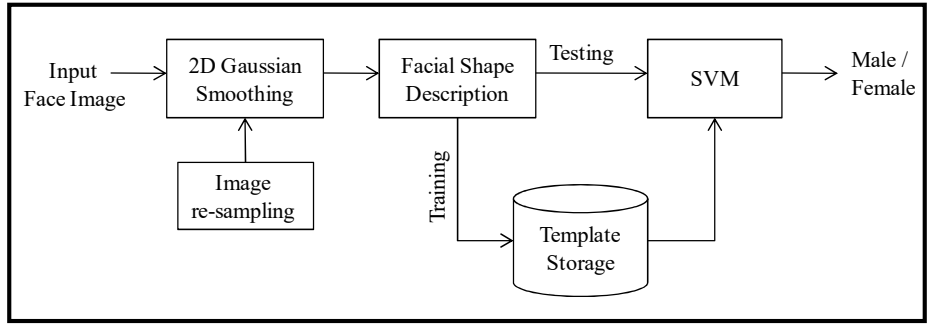

*Figure 1.* **Operations required for performing gender recognition**

In the pre-processing phase, the face image is enlarged via image re-sampling based on bi-cubic interpolation and noise filtering using 2D-Gaussian kernel. The feature extraction phase combines Laplacian edge detector with PHOG descriptor, for extracting local shape description of the face image. In the classification phase, we utilized Support Vector Machine (SVM) for classifying the genders.

**Pre-processing**

The images utilized in this paper are already cropped to equal dimension by the authors of LFW database (Huang, Mattar, Berg, & Learned-Miller, 2007; Sanderson & Lovell, 2009). Therefore, a two-step pre-processing is considered in this experiment. Initially, the images are re-sampled to enlarge the spatial resolution to 300 by 300 pixels using bi-cubic interpolation. Image re-sampling is mainly considered in this experiment in order to enhance the facial shape feature representation, as the enlarged image depicts more detailed structure of the face. Besides, by using bi-cubic interpolation, the structure of the image can still be preserved (Wittman, 2005). Then, a 2D Gaussian filter is applied to smooth the normalized image.

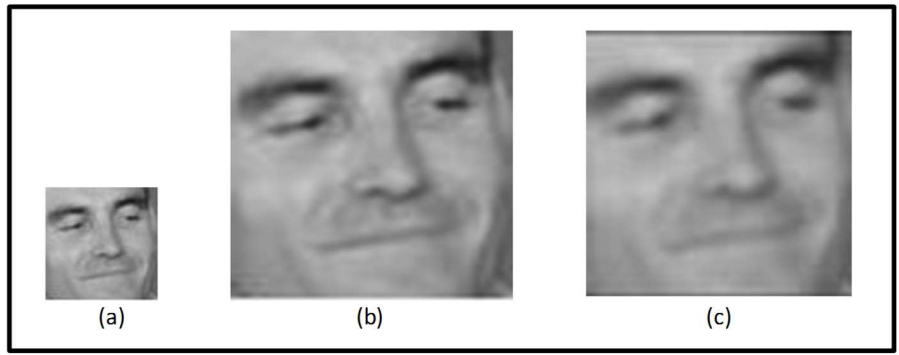

*Figure 2.* **Face pre-processing stages (a) Original image of 64 x64 pixels.
(b) Re-sampled image with bi-cubic interpolation (c) Smoothed image with 2D Gaussian.**

Performance of Gaussian filter is computationally efficient since it performs consistently in different directions. The function of a 2-D Gaussian filter can be represented using the following mathematical expression:

$$G(x,y) = \frac{1}{\sqrt{2\pi\sigma^2}} e^{-\frac{x^2+y^2}{2\sigma^2}} \tag{1}$$

Where x and y are the coordinate of the pixel in the image and σ is the variance. The variance denotes the tightness of the width of the Gaussian function (Gonzalez & Woods, 2002).

**Feature Extraction**

From the output of image pre-processing, the next step involves extracting important feature that can differentiate between the two genders. The aim in this paper is to offer the description of the face using local feature descriptor. As such, we consider shape representation as an important feature, since its physical description provides indications that can be used to differentiate between genders. Based on image processing techniques, shape can be extracted in the form of edges or contours, but it can be quantified using feature descriptors. To achieve this, the shape representation algorithm described by (Bosch et al., 2007) is adopted. With regard to this, we utilized the Laplacian operator for edge detection rather than the Canny edge detector that was used in the original implementation of the algorithm. The reason for using Laplacian operator is because it is an isotropic filter. It can detect edge information equally without bias on the orientation or direction of the edges. In order to detect edges, the spatial kernel of the Laplacian operator is convolved with the image.

The output of this convolution between the image and the Laplacian operator is normally grayish edge lines. It highlights the regions of swift change in pixel intensity and other discontinuities, while suppressing the regions where there are gradual changes in pixel intensity, which are superimposed on an empty dark background (Gonzalez & Woods, 2002). The result of applying the Laplacian operator is shown in Figure 3. PHOG descriptor is inspired by the concept of Histogram of Oriented Gradient proposed by Dalal and Triggs (Dalal & Triggs, 2005). PHOG can be considered as a combination of several HOGs with a minor variation in its computation.

Computing the PHOG descriptor involves the following steps.

1. Divide the image into regions which are referred to as pyramid levels (see Bosch et al., 2007).
2. Use second order derivative operator based on Laplacian operator to generate edges in different directions in the image as shown in Figure 3(a).
3. Apply sobel operator to generate image gradients, $Gr_x$ and $Gr_y$ in horizontal and vertical directions.
4. Compute the gradient magnitude $\nabla I$ and orientation $\theta$ respectively using equations 2 and 3:

$$\nabla I = \sqrt{Gr_x^2 + Gr_y^2} \qquad (2)$$

$$\theta = tan^{-1}(Gr_y/Gr_x) \qquad (3)$$

5. Map the orientations to a specific range using equation 4, depending on the number of bins used for the histogram.

$$O_H = \left(\frac{\theta}{\frac{A}{n-bins}}\right) \qquad (4)$$

Where, $A$ is the predefined angle for mapping the edge orientation histograms.

6. Compute a weighted summation of gradient magnitude values at a dense spatial grid based on the direction of the edge orientation element positioned on it (Bosch et al., 2007).

The weighted votes are accumulated into cells of orientation bins over local spatial regions (Bosch et al., 2007). The resulting edge orientation and gradient histograms at each spatial pyramid are then concatenated into a feature descriptor which gives a more compact shape descriptor, as shown in Figure 3(d). The feature length of the descriptor can be varied depending on the application domain by reducing or increasing the number of pyramid levels, $L$, and number of bins for the binning local histogram $H$.

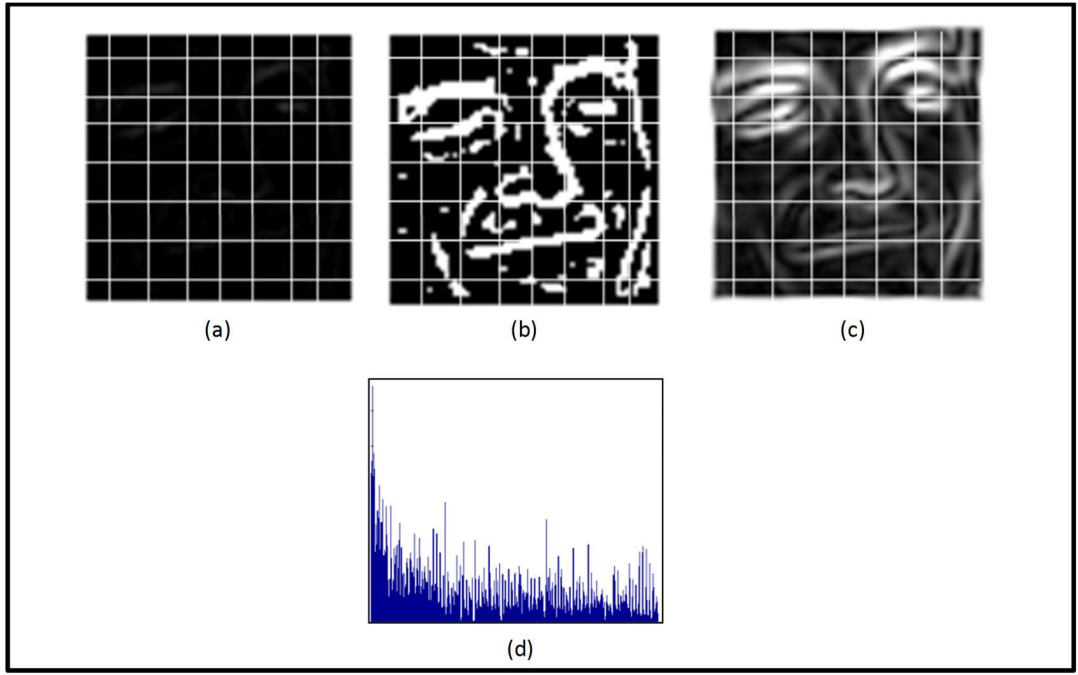

*Figure 3.* Facial shape feature representation process. (a) Output of Laplacian edge detection (b) Highlighting the edge regions  (c) Image gradient magnitude  (d) Laplacian PHOG descriptor

**Classification**

Considering the fact that, the face features have been extracted, thus, the last operation to be performed is classification, where the system learns to classify the features into their respective classes. Gender recognition is considered as a binary classification problem involving only two classes (either male or female). Support Vector Machine ( Cortes & Vapnik, 1995; Zaki et al., 2002) is a suitable classifier for two class learning problem such as recognition of genders from human faces. SVM is a popular supervised learning algorithm in the domain of machine learning for performing classification, regression, and outlier detection. SVM is simply a structural risk minimization algorithm based on statistical learning theory. The main concept is aimed at finding an optimal separating hyperplane that sufficiently separates the data. Detailed mathematical theorems on SVM can be found in (Cortes & Vapnik, 1995). In this paper, we consider the soft margin SVM, usually adopted when the input data space is not linearly separable. This is because there are some situations whereby the datasets are not linearly separable, therefore the input space in which the data points are not linearly separable can be mapped or transformed to a higher dimensional feature space through a mapping function $\emptyset(x_i)$, where the data can be linearly separable. In this case, the soft margin hyperplane can be found by incurring slack variable $\xi_i$ derived from allowing some errors during training. The problem of finding the hyperplane with minimizing the training errors can be expressed as follows (Mustaffa et al., 2013):

$$\text{Minimize:} \quad \frac{1}{2}\mathbf{w}^T\mathbf{w} + C\sum_{i=1}^{M}\xi_i$$

$$subject\ to \quad y_i(\mathbf{w}\cdot\emptyset(\mathbf{x_i}) + b) \geq 1 - \xi_i, \quad i = 1,\dots M \quad (5)$$

$$\xi_i \geq 0, \quad i = 1,\dots M$$

Where $w$ is the normal to the optimal separating hyperplane and $b$ is the bias term. The user defined parameter $C$ in the equation is the cost parameter which determines the tradeoff between the training error and distance of hyperplane. It means that, we allow some errors during training for the classifier to find generalization on new test data. Training this kind of SVM is equivalent to solving the dual optimization problem and its Lagragian formulation $L_p$ can be expressed as follows:

$$\text{Maximize: } L_p(\alpha_i) = \frac{1}{2}\sum_{i=1}^{M}\sum_{j=1}^{M} \alpha_i \alpha_j\, y_i y_j\, \emptyset(\mathbf{x_i}) \cdot \emptyset(\mathbf{x_j}) - \sum_{i=1}^{M} \alpha_i$$

$$\text{subject to: } \sum_{i=1}^{M} y_i \alpha_i = 0 \tag{6}$$

$$0 \leq \alpha_i \leq C, \quad i = 1, \dots, M$$

Where $\alpha_i$ is a Langrage multiplier corresponding to the training sample, $x_i$. The set of $\alpha$, that are non-zero are regarded as the support vectors. Further, the inner products $\emptyset(x_i) \cdot \emptyset(x_j)$ can be replaced with kernel function $K(x_i\, x_j)$ and different type of kernels can be utilized for different non-linear problem. In this paper three kernel functions are considered, namely Linear, Radial Basis Function (RBF), and Polynomial kernel.

## EXPERIMENTAL RESULTS

An evaluation of the proposed technique is performed on cropped Labeled Faces in the Wild (LFW) database (Huang et al., 2007; Sanderson & Lovell, 2009). The data was specifically compiled with the intentions of studying the problem of face recognition from unconstrained images. The composition of the dataset provides several photometric challenges that represent real life situations such as misalignment, scale, pose, illumination, occlusion, and expression variations. It contains 13,233 facial images of 5,749 individuals with each image having a spatial resolution of 64 x 64 pixels. The total number of male face images is 10256 and female face images of 2977. Due to this imbalance between male and female images, there tends to be variations in the number of samples available for each subject. A certain number of individuals have only one sample, while some others possess approximately 150 sample images. Hence, we annotated ground truth for each image in the dataset based on our subjective perception. Based on this, we selected the images to be used for the experiment using the following criteria:

- Quality of the image
- Clarity of gender prediction by human annotator
- Frontal or near frontal view profile

Eventually, a total of 2,759 face images are selected, which is composed of 1,679 different male faces and 1,080 different female faces. Some sample images are shown in Figure 4.

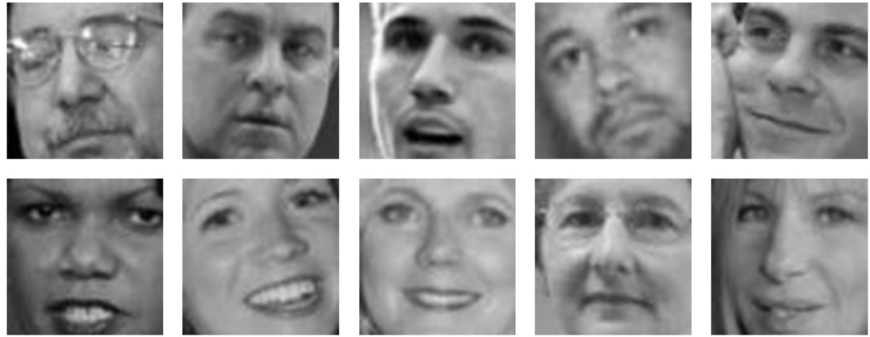

*Figure 4.* Some sample face images from cropped LFW dataset.

To perform classification, the data is randomly partitioned into 2 equal folds of training and testing set for the two classes. The input data undergo the stage of preprocessing with bi-cubic interpolation and Gaussian noise removal. Afterwards, the normalized image is input to the feature extraction stage involving Laplacian PHOG shape description of the face. For the feature descriptor, the pyramid level $L$ is varied between 2 and 3, while the binning histogram $H$ is varied between 8 and 16, as shown in Table 2. In addition, SVM with three different kernel functions are experimented on each feature descriptor, the results are presented in Table 2. The parameter $C$ for the three kernels is varied from $2^{-5}$ to $2^{10}$, the gamma $\gamma$ for the RBF kernel is varied from $2^{-10}$ to $2^{10}$. Finally, the recognition rate of the system is presented by calculating the percentage of True Positive (TP) and True Negative (TN) over the entire population, using equation 7.

$$Recognition\ Rate = \frac{TP+TN}{Total\ Population} \quad (7)$$

**Table 2**

**Results of gender classification with kernel SVM**

| Descriptor Parameters | Feature Length | Kernel Function | Recognition Rate (%) | Best Parameter $C$ | $\gamma$ |
|---|---|---|---|---|---|
| L = 2, H = 8 | 168 | Linear | 79.4 | $2^{-2}$ | |
| | | Polynomial | 82.5 | $2^{-3}$ | |
| | | **RBF** | **83.6** | **$2^{3}$** | **$2^{9}$** |
| L = 2, H = 16 | 336 | Linear | 77 | $2^{5}$ | |
| | | Polynomial | 82.6 | $2^{3}$ | |
| | | **RBF** | **84.2** | **$2^{4}$** | **$2^{10}$** |
| L = 3, H = 8 | 680 | Linear | 82.7 | $2^{-4}$ | |
| | | Polynomial | 85.6 | $2^{1}$ | |
| | | **RBF** | **86.8** | **$2^{3}$** | **$2^{9}$** |
| **L = 3, H = 16** | **1360** | Linear | 81.9 | $2^{-5}$ | |
| | | Polynomial | 88.1 | $2^{7}$ | |
| | | **RBF** | **88.5** | **$2^{2}$** | **$2^{10}$** |

The parameter setting $L = 2$, $H = 8$ extracts a feature vector with length of 168, accumulated over 2 pyramid levels and binned with histogram of *8* bins. An initial attempt on linear kernel with SVM on the feature descriptor yielded a recognition rate of 79.4%, Polynomial kernel attained a recognition rate of 82.5%, while RBF provided the highest recognition rate of 83.6%. Increasing the binning histogram to *16* increases the length of the feature to 336, and the top performance of 84.2% is attained with the RBF kernel. At a more dense pyramid level $L = 3$, the face information being accumulate increases, resulting in an improvement in the recognition rate to 86.8% with RBF kernel. Finally, an increase in the binning histogram to $H = 16$ further increases the feature length to 1360 and the best result of 88.5% is attained with RBF kernel. We inferred in this experiment that increasing the pyramid level has more effect on gender recognition result than increasing the binning histogram. This is because, at the same pyramid level, changing the number of binning histogram results into a longer feature vector with redundant and sparse components in the vector. However, at new pyramid level, several new information are encoded into the feature vector even though the size has increased. In addition, the RBF kernel is found to be more suitable for transforming the data to high dimensional space, as it improves the performance of the system.

## DISCUSSION AND CONCLUSION

This paper proposed an approach for gender recognition from unconstrained face images using facial shape description and kernel based SVM. The technique was implemented using sample images from the cropped LFW dataset. The images utilized possess some level of challenges in terms of head pose variation, face expression, illumination condition, occlusion, and low image resolution. The proposed method makes use of only a single sample image for each subject, in contrary to the previous techniques where multiple sample images were required for each individual. Our method adopts existing shape descriptor based on spatial pyramid of orientation and gradient histogram, with support vector machine for classification. The main aim of this paper was to determine the most suitable functions and parameter combinations for performing gender recognition on unconstrained face image. The experimental results indicated that PHOG descriptor with pyramid level L = 3 and binning histogram H =16, combined with RBF kernel SVM can attain an optimum performance of 88.5%.

## REFERENCES

Arigbabu, O. A., Ahmad, S. M. S., Adnan, W. A. W., & Yussof, S. (2014). Recent advances in facial soft biometrics. *The Visual Computer*, 1–13. doi:10.1007/s00371-014-0990-x

Arigbabu, O. A., Ahmad, S. M. S., Adnan, W. A. W., Yussof, S., Iranmanesh, V., & Malallah, F. L. (2014a). Estimating Body Related Soft Biometric Traits in Video Frames. *The Scientific World Journal*, *2014*, 1–13.

Arigbabu, O. A., Ahmad, S. M. S., Adnan, W. A. W., Yussof, S., Iranmanesh, V., & Malallah, F. L. (2014b). Gender recognition on real world faces based on shape representation and neural network. In *Proceedings of 2014 IEEE International Conference on Computer and Information Sciences (ICCOINS)* (pp. 1–5).

Bekios-Calfa, J. (2013). Alignment-Free Gender Recognition in the Wild. In *Pattern Recognition and and Image Analysis* (Vol. 1, pp. 382–389).

Bekios-Calfa, J., Buenaposada, J. M., & Baumela, L. (2014). Robust gender recognition by exploiting facial attributes dependencies. *Pattern Recognition Letters*, *36*, 228–234.

Bosch, A., Zisserman, A., & Munoz, X. (2007). Representing shape with a spatial pyramid kernel. In *Proceedings of the 6th ACM international conference on Image and video retrieval* (pp. 401–408).

Chu, W.-S., Huang, C.-R., & Chen, C.-S. (2013). Gender classification from unaligned facial images using support subspaces. *Information Sciences*, *221*, 98–109.

Cortes, C., & Vapnik, V. (1995). Support-vector networks. *Machine Learning*, *297*, 273–297.

Dago-casas, P., Gonzalez-Jimenez, D., Yu, L. L., & Alba-Castro, J. L. (2011). Single- and Cross- Database Benchmarks for Gender Classification Under Unconstrained Settings. In *Proceedings of IEEE International Conference on Computer Vision Workshops* (pp. 2152–2159).

Dalal, N., & Triggs, B. (2005). Histograms of Oriented Gradients for Human Detection. In *IEEE Computer Society Conference on Computer Vision and Pattern Recognition* (Vol. 1, pp. 886 – 893).

Farinella, G., & Dugelay, J. (2012). Demographic classification: Do gender and ethnicity affect each other? In *Proceedings of IEEE Intenational Conference on Informatics, Electronics & Vision* (pp. 383–390).

Golomb, B., Lawrence, D., & Sejnowski, T. (1990). SEXNET: A Neural Network Identifies Sex From Human Faces. In *Proceedings of 1990 conference on Advances in neural information processing systems 3 (NIP-3). Richard P. Lippmann, John E. Moody, and David S. Touretzky (Eds.). Morgan Kaufmann Publishers Inc., San Francisco, CA, USA.* (pp. 572–577).

Gonzalez, R. C., & Woods, R. E. (2002). Digital Image Processing (3rd Edition). In *Prentice-Hall, Inc*.

Huang, G., Mattar, M., Berg, T., & Learned-Miller, E. (2007). Labeled faces in the wild: A database for studying face recognition in unconstrained environments. *University of Massachusetts, Amherst, Technical Report*, 1–14.

Khan, S. A., Nazir, M., Akram, S., & Riaz, N. (2011). Gender classification using image processing techniques: A survey. In *Proceedings of the 2011 IEEE 14th International Multitopic Conference (INMIC)* (pp. 25–30).

Makinen, E., & Raisamo, R. (2008). Evaluation of gender classification methods with automatically detected and aligned faces. *IEEE Transactions on Pattern Analysis and Machine Intelligence*, *30*(3), 541–547.

Mayo, M., & Zhang, E. (2008). Improving face gender classification by adding deliberately misaligned faces to the training data. In *IEEE 3rd International Conference on Image and Vision Computing* (pp. 1–5).

Mustaffa, Z., Yusof, Y., & Kamaruddin, S. (2013). Enhanced ABC-LSSVM for Energy Fuel Price Prediction. *Journal of ICT*, *12*, 73–101.

Rai, P., & Khanna, P. (2012). Gender Classification Techniques: A Review. *Advances in Computer Science, Engineering & Applications*, *166*, 51–59.

Sanderson, C., & Lovell, B. (2009). Multi-region probabilistic histograms for robust and scalable identity inference. *Advances in Biometrics*, *5558*, 199–208.

Shan, C. (2012). Learning local binary patterns for gender classification on real-world face images. *Pattern Recognition Letters*, *33*(4), 431–437.

Tamura, S., Hideo, K., & Mitsumoto, H. (1996). Male / Female Identification from 8 x 6 Very Low Resolution Face Images by Neural Network. *Pattern Recognition*, *29*(2), 331–335.

Wittman, T. (2005). Mathematical techniques for image interpolation. *Department of Mathematics University of Minnesota*, 1–33.

Yang, M., & Moghaddam, B. (2000). Support vector machines for visual gender classification. In *Proceedings of the 15th International Conference on Pattern Recognition* (pp. 1115–1118).

Zaki, N., Deris, S., & Chin, K. (2002). Extending the decomposition algorithm for support vector machines training. *Journal of ICT*, *1*(1), 17–29.